\PassOptionsToPackage{table,dvipsnames}{xcolor}
\documentclass[letterpaper, 10 pt, conference]{ieeeconf}
\usepackage{icra}

\IEEEoverridecommandlockouts                              

\definecolor{somegray}{rgb}{0.5, 0.5, 0.5}
\newcommand{\darkgrayed}[1]{\textcolor{somegray}{#1}}
\makeatletter
\newcommand*\titleheader[1]{\gdef\@titleheader{#1}}
\AtBeginDocument{%
  \let\st@red@title\@title
  \def\@title{%
    \vskip-3.5em
    \bgroup\normalfont\large\centering\@titleheader\par\egroup
    \vskip1.5em\st@red@title}
}
\makeatother
\titleheader{\darkgrayed{This paper has been accepted for publication at the\\IEEE International Conference on Robotics and Automation, Atlanta, 2025
\copyright IEEE}}
\title{\LARGE \bf
Environment as Policy: Learning to Race in Unseen Tracks
}

\author{Hongze Wang*, Jiaxu Xing*, Nico Messikommer, and Davide Scaramuzza
\thanks{$^{*}$ equal contribution.
    The authors are with the Robotics and Perception Group, Department of Informatics, University of Zurich, and Department of Neuroinformatics, University of Zurich and ETH Zurich, Switzerland (\protect\url{http://rpg.ifi.uzh.ch}). This work was supported by the European Union’s Horizon Europe Research and Innovation Programme under grant agreement No. 101120732 (AUTOASSESS) and the European Research Council (ERC) under grant agreement No. 864042 (AGILEFLIGHT).}
}
\begin{document}
\makeatletter
\g@addto@macro\@maketitle{
  \captionsetup{type=figure}\setcounter{figure}{0}
  \def\mycolspace{1.2mm}
  \centering
  \vspace{1mm}
  \tikzset{font=\small}
 \centering
	\begin{tikzpicture}[>=latex,scale=0.85, every node/.style={scale=0.8}, every text node part/.style={align=center}]
		\tikzset{every node/.style={minimum width=0.6cm,minimum height=0.5cm}}
    \node [inner sep=0pt, outer sep=0pt] (img1) at (0, 0) 
    {\includegraphics[width=\linewidth]{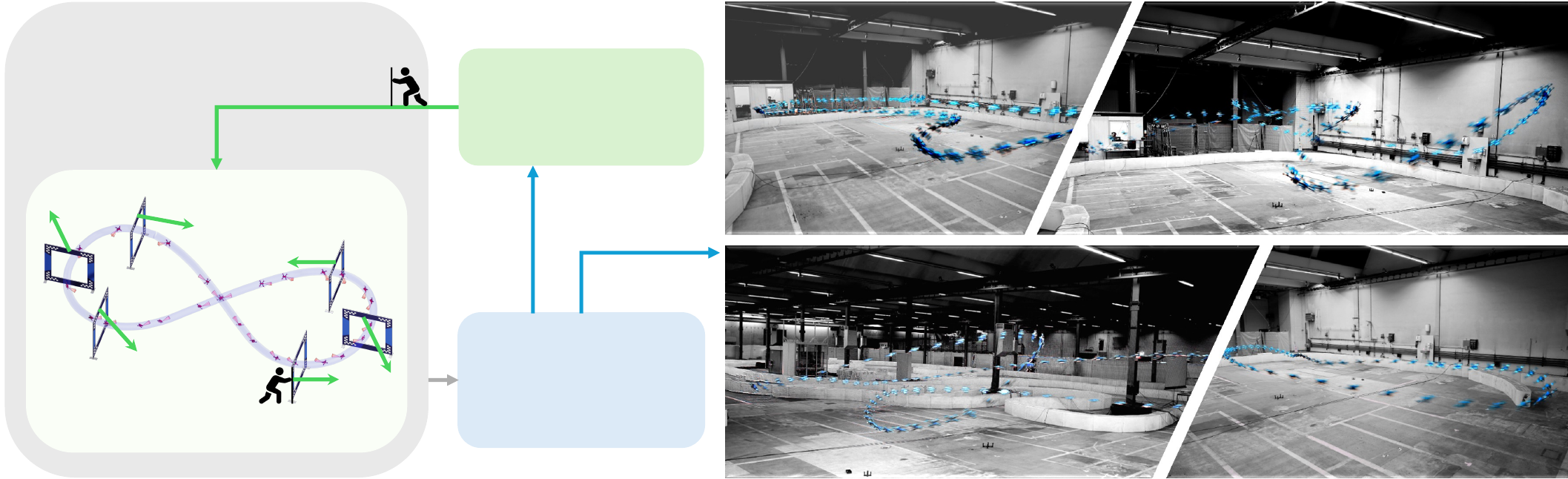}};
    \node (training) at (-7.4, 2.6) {\textbf{Policy Training}};
    \node (sac) at (-2.6, 1.75) {\textit{Environment} \\ Policy (SAC)};
    \node (ppo) at (-2.6, -1.85) {\textit{Racing} \\ Policy (PPO)};
    \node (deployment) at (5, 3.5) {\textbf{Real World Deployment}};
    \node [rotate=90, text=gray] (rollout) at (-3.7, -0.14) {\small Rollout};
     \node (gate)[align=right, text=gray] at (-9, 1.48) {\small Environment\\ \small Shaping};
    \node (gate)[align=right, color=ForestGreen] at (-8.95, -2.42) {\small Track Layout \\ \small Changes};
    \node[draw, fill=gray!10, rectangle] at (9.29, 2.85) {\small (b) 3D Big S};
    \node[draw, fill=gray!10, rectangle] at (0.37, 2.85) {\small (a) 2D Big S};
    \node[draw, fill=gray!10, rectangle] at (0.05, -2.85) {\small (c) Twist};
    \node[draw, fill=gray!10, rectangle] at (9.48, -2.86) {\small (d) Kidney};
	\end{tikzpicture}
    \vspace{0.5ex}
	\captionof{figure}{In this work, we introduce an environment-shaping framework to improve the generalization of RL drone racing agents to diverse and unseen tracks without retraining. We use a Soft Actor-Critic (SAC) policy to adaptively shape the track layouts by generating difficult but achievable race tracks based on the racing agent's actual performance. The resulting \textit{single} racing policy can fly in various unseen and challenging race tracks in the real world with competitive lap times.}
\vspace{-0.2cm}
\label{fig:overview}
}
\makeatother
\makeatletter
{\let\newpage\relax\maketitle}

\begin{abstract}
Reinforcement learning (RL) has achieved outstanding success in complex robot control tasks, such as drone racing, where the RL agents have outperformed human champions in a known racing track.
However, these agents fail in unseen track configurations, always requiring complete retraining when presented with new track layouts.
This work aims to develop RL agents that generalize effectively to novel track configurations without retraining.
The na\"ive solution of training directly on a diverse set of track layouts can overburden the agent, resulting in suboptimal policy learning as the increased complexity of the environment impairs the agent's ability to learn to fly.
To enhance the generalizability of the RL agent, we propose an adaptive environment-shaping framework that dynamically adjusts the training environment based on the agent's performance. 
We achieve this by leveraging a secondary RL policy to design environments that strike a balance between being challenging and achievable, allowing the agent to adapt and improve progressively.
Using our adaptive environment shaping, one single racing policy efficiently learns to race in diverse challenging tracks.
Experimental results validated in both simulation and the real world show that our method enables drones to successfully fly complex and unseen race tracks, outperforming existing environment-shaping techniques.
Website: \protect\url{http://rpg.ifi.uzh.ch/env_as_policy}.
\end{abstract}
\section{Introduction}

Reinforcement learning (RL) involves agents learning through trial and error by interacting with a pre-defined environment and maximizing the rewards based on these interactions.
It has proven highly effective in various robotic control applications, demonstrating remarkable task performance across scenarios like dexterous manipulation~\cite{andrychowicz2020learning, kalashnikov2018scalable, kalashnikov2022scaling, aljalbout2024role}, quadrupedal locomotion~\cite{lee2020learning, kumar2021rma}, and agile quadrotor flight~\cite{song2021autonomous, messikommer2024contrastive, xing2024contrastive, krinner2025accelerating, romero2024actor2}. 
Previous research has shown that RL can even outperform human champions in the task of drone racing~\cite{kaufmann23champion, Song23Reaching}, where an RL agent flies a quadcopter drone through a complex track at high speed, requiring quick decision-making and precise control.

However, while RL agents excel within the specific distributions they are trained on, they struggle with out-of-distribution configurations and may require retraining from scratch for even minor configuration changes~\cite{DR}.
To improve adaptability and generalization, extending the RL framework to train and perform across a broader range of distributions is essential, enabling agents to handle more diverse and dynamic environments effectively.
However, training directly on a wide distribution of scenarios can significantly degrade the efficiency and effectiveness of the learning process~\cite{bengio2009curriculum}.
As the distribution broadens, the complexity of policy exploration increases, making it harder for the agent to identify and implement effective solutions.

Domain randomization is a commonly used technique to improve the learning capability of RL agents across a broader range of tasks~\cite{DR, DR2, openai2019learning, openai2019solving}.
This approach varies the parameters of the training environment to expose the agent to a wide variety of potential deployment scenarios.
By learning in diverse conditions, the agent develops robust policies less likely to overfit the specific characteristics of a single environment.
Domain randomization is often combined with curriculum learning, where the complexity and variability of training scenarios are incrementally increased~\cite{openai2019solving}.
By starting with less challenging environments and gradually introducing greater variability, the agent learns to perform well in a wide set of scenarios without becoming overwhelmed during the early stages of training.
However, curriculum learning often depends on manual design, introducing human biases.
This reliance on fixed, non-adaptive progressions can limit training diversity and restrict the agent’s ability to generalize to new, real-world situations~\cite{kong2021adaptive}.
These effects are especially prominent in the challenging task of drone racing, where the platform's high agility and the need for rapid decision-making amplify the difficulties.
Therefore, the challenge of agile drone racing on an unseen race track in real-world conditions remains largely unexplored.

To address this problem, we propose an automatic adaptive environment-shaping framework to enhance the agent's generalization ability to fly in unseen race tracks. 
This framework dynamically adjusts the environment curriculum based on the agent's learning progress (Fig.~\ref{fig:overview}).
The key idea is to create consistently challenging yet attainable environments, avoiding tracks that are too easy or overly difficult, which could hinder learning.
We achieve this by leveraging a secondary RL agent to design environments that maintain a balanced difficulty level, allowing the agent to adapt and improve progressively.
By automating the environment shaping, the system can more precisely tailor the learning environment to the agent’s performance, reducing human bias and significantly enhancing the agent's ability to generalize across various unseen track configurations.
In experiments validated both in simulation and the real world, we demonstrate for the first time a drone racing policy that can race in different unseen tracks.
We show that our approach outperforms the existing environment-shaping approaches using the same number of actions and enables the racing policy to generalize to a diverse set of unseen and complicated tracks.

\section{Related Works}
\textbf{Reinforcement Learning for Robotics}
Reinforcement Learning (RL) has been extensively applied in robotics to facilitate continuous control in complex and dynamic environments.
Recent studies have achieved remarkable success in various robotic domains.
These include quadrotor control~\cite{kaufmann23champion, Song23Reaching, hwangbo2017control, xing2024contrastive, xing2024bootstrapping, messikommer2024student}, legged locomotion~\cite{hwangbo2019learning, kumar2021rma}, and manipulation~\cite{fu2023deep, gu2017deep}, showcasing significant advancements in the field.

\textbf{Autonomous Drone Racing}
Autonomous drone racing is an emerging field in which drones must navigate through predefined waypoints in the shortest possible time. 
Success in this field requires integrating advanced hardware and sophisticated algorithms capable of perceiving the environment, planning optimal paths, and executing actions in real-time~\cite{hanover2024autonomous}.
\cite{Song23Reaching} demonstrated that RL controllers can outperform traditional optimization-based approaches, such as those proposed in~\cite{romero2022model, romero2024actor}, for autonomous drone racing. 
This advantage arises because RL can effectively handle highly non-linear dynamical systems and complex objectives, which pose significant challenges for conventional optimization methods like Model Predictive Control (MPC)~\cite{falanga2018pampc, xing2023autonomous}.
Recent studies have shown that policies trained with deep reinforcement learning (DRL) can surpass even human world champions, employing vision-based state estimation for the RL controller. 
Furthermore, ~\cite{geles2024demonstrating, xing2024bootstrapping} have demonstrated using RL to learn drone racing from image inputs directly.

\textbf{Environment Shaping for Reinforcement Learning}
Environment shaping has been widely applied in reinforcement learning by designing a distribution of environments that progressively improves the agent's performance.
Previous works like~\cite{portelas2019, parkerholder2023, jiang2022, dennis2021} on environment shaping have been successful in simulation tasks like Bipedal Walker and maze games, where environments are easily parameterized for manual curriculum design. 
However, this process becomes more challenging for more complex real-world robotic tasks. Domain randomization is a common solution, where environments are sampled from a predefined range to enable robust task execution. 
For instance, ~\cite{openai2019solving, openai2019learning} used domain randomization to achieve dexterous manipulation in the real world. 
Additionally, ~\cite{lee2020learning} proposed an adaptive terrain curriculum using a particle filter to sample environment parameters, enabling quadrupedal locomotion across various terrains in real-world scenarios.

\section{Methodology}
\begin{figure*}[ht]
\centering
\includegraphics[width=\linewidth]{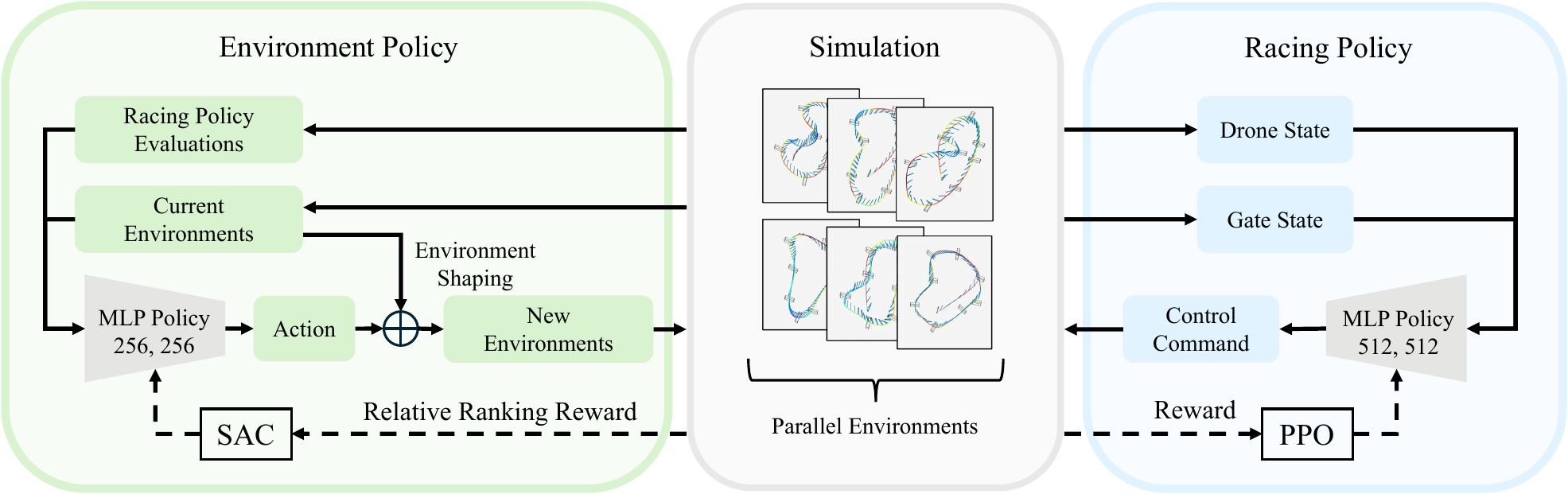}
\caption{Overview of the proposed method. In every N iteration, the environment policy \textit{(left)} takes as input the information of the racing policy evaluations and the current environments. It generates actions to adjust the gate layouts independently for each parallel environment. The racing policy (right) utilizes the information about drone and gate states from these simulation environments to learn time-optimal drone racing strategies through an MLP.}
\label{fig:pipeline}
\vspace{-0.6cm}
\end{figure*} 
To enable agile flight through unseen tracks, our approach introduces an \textit{Environment Policy} that acts as a separate agent, guiding the racing policy's training across varied tracks. 
Both policies are employed in an alternating fashion during training: the environment policy creates tracks tailored to the drone agent's learning progress, enhancing the racing policy's robustness and speed.
An overview of the method is visualized in Fig.~\ref{fig:pipeline}.
In the following, we introduce the racing policy in Sec.~\ref{sec:agent_policy} and describe the proposed environment policy in Sec.~\ref{sec:environment_policy}.
\subsection{Racing Policy Training}
\label{sec:agent_policy}
The autonomous racing task can be framed as an optimization problem, where the objective is to minimize the time it takes for an agile quadrotor to pass through a predefined sequence of gates~\cite{hanover2024autonomous}, as illustrated in Fig.~\ref{fig:track}.
In this task, we define the observations as $\bm{o}_{\mathrm{racing}} = \left[\tilde{\bm{R}},  \bm{v}, \bm{\omega}, a_\text{prev}, 
\delta\bm{p}_1, \delta\bm{p}_2\right]$, 
where $\tilde{\bm{R}} \in \mathbb{R}^6$ is a vector comprising the first two columns of $\bm{R}_{\wfr\bfr}$~\cite{zhou2019continuity},
$\bm{v} \in \mathbb{R}^3$ and $\bm{\omega} \in \mathbb{R}^3$ denote the linear and angular velocity of the drone, 
$a_\text{prev}$ represents the previous action from the actor policy,
and $\delta\bm{p}_1, \delta\bm{p}_2 \in \mathbb{R}^{12}$ represent the relative difference in position of the four next gate corners ($4\times 3$) in the world frame.
Here, $\delta\bm{p}_1$ represents the difference of the four corners of the next gate to pass between the current quadrotor position, and $\delta\bm{p}_2$ represents the positional difference of the corners between the next gate to pass and the gate after the next gate to pass on the race track.
The total reward at time $t$, denoted as $r_t$, consists of several components:
\begin{equation}
    r_t^{\mathrm{racing}} = r_t^{\mathrm{prog}} + r_t^{\mathrm{act}} + r_t^{\mathrm{br}} + r_t^{\mathrm{pass}} + r_t^{\mathrm{crash}},
\end{equation}
where $r_t^{\mathrm{prog}}$ represents progress toward passing the next gate~\cite{yunlong2021racing}, 
$r_t^{\mathrm{act}}$ penalizes changes in actions from the previous time step, 
$r_t^{\mathrm{br}}$ discourages high body rates to ensure a stable flying behavior, 
$r_t^{\mathrm{pass}}$ is a binary reward for successfully passing the next gate, 
and $r_t^{\mathrm{crash}}$ is a binary penalty applied when a collision occurs, which also terminates the episode.
The reward components are formulated as follows
\begin{equation}
    \begin{aligned}
        r_t^{\mathrm{prog}} & = \alpha_1(d_{\mathrm{Gate}}(t-1) - d_{\mathrm{Gate}}(t)), \\
        r_t^{\mathrm{act}} & = \alpha_2 \lVert\bm{u}_t - \bm{u}_{t-1}\rVert, \\
        r_t^{\mathrm{br}} & = \alpha_3\lVert\bm{\omega}_{\mathcal{B},t}\rVert, \\
        r_t^{\mathrm{pass}} & =  \alpha_4  \quad\text{if robot passes the next gate}, \\
        r_t^{\mathrm{crash}} & = \alpha_5 \quad\text{if robot crashes (gates, ground)}. 
    \end{aligned}
    \label{eq:reward}
\end{equation}

\subsection{Environment Policy Training}
\label{sec:environment_policy}
The track layout used for training the racing policy plays a critical role in shaping its flying capabilities while also impacting the overall training stability.
If the tracks are too difficult, the agent will struggle to extract meaningful learning signals.
Conversely, simple tracks fail to challenge the agent, limiting its ability to generalize to more complex environments.
Thus, to ensure effective learning, the track difficulty must be \textit{continuously adapted to the current capabilities of the agent}.
To achieve this, we introduce a learned environment policy $\pi_{\mathrm{env}}$ that dynamically adjusts the race tracks. 
This adaptive approach allows the agent to consistently gather relevant and progressive learning experiences, optimizing its training stability and performance in diverse race tracks.
\subsubsection{MDP Formulation}
In our racing scenario, the states of the environment are represented by the position $\bm{p}$ and orientation $\bm{R}$ of each individual gate, 
such that the full gate state vector is given by $\bm{s}_\mathrm{gates} = [\bm{p}_1, \bm{R}_1, \dots, \bm{p}_N, \bm{R}_N]$, where $N$ is the total number of gates in the environment.
The environment policy leverages the observation $\bm{o}_{\mathrm{env}} = [\bm{s}_\mathrm{gates}, \bm{e}_\mathrm{racing}]$,
where $s_\mathrm{gates}$ represents the current states of all gates, and $\bm{e}_\mathrm{racing}$ indicates the performance of the drone agent achieved in the track corresponding to the environment state $s_\mathrm{gates}$.
The evaluation performance vector $\bm{e}_\mathrm{racing} = [\bm{e}_1, \dots, \bm{e}_N]$ contains the gate-passing error for each of the $N$ gates, 
where the error $\bm{e}_i$ is computed as the distance between the drone's position and the center of gate $i$ at the moment the drone passes through the gate.
The environment policy outputs actions $a= [\Delta \bm{p}_{\mathrm{gates}}, \Delta \mathrm{yaw}_{\mathrm{gates}}]$, where $\Delta \bm{p}_{\mathrm{gates}}$ specifies changes in gate positions, and $\Delta \text{yaw}_{\text{gates}}$ indicates changes in the yaw angles of the gates, both in the world frame coordinates. 
Hence, the transition dynamics $\mathbb{P}$ is naturally $s^t_\mathrm{gates} = s^{t-1}_\mathrm{gates} + a^{t-1}$.
As training progresses, the training track layout reflects accumulated changes from the initial track, naturally producing various tracks.
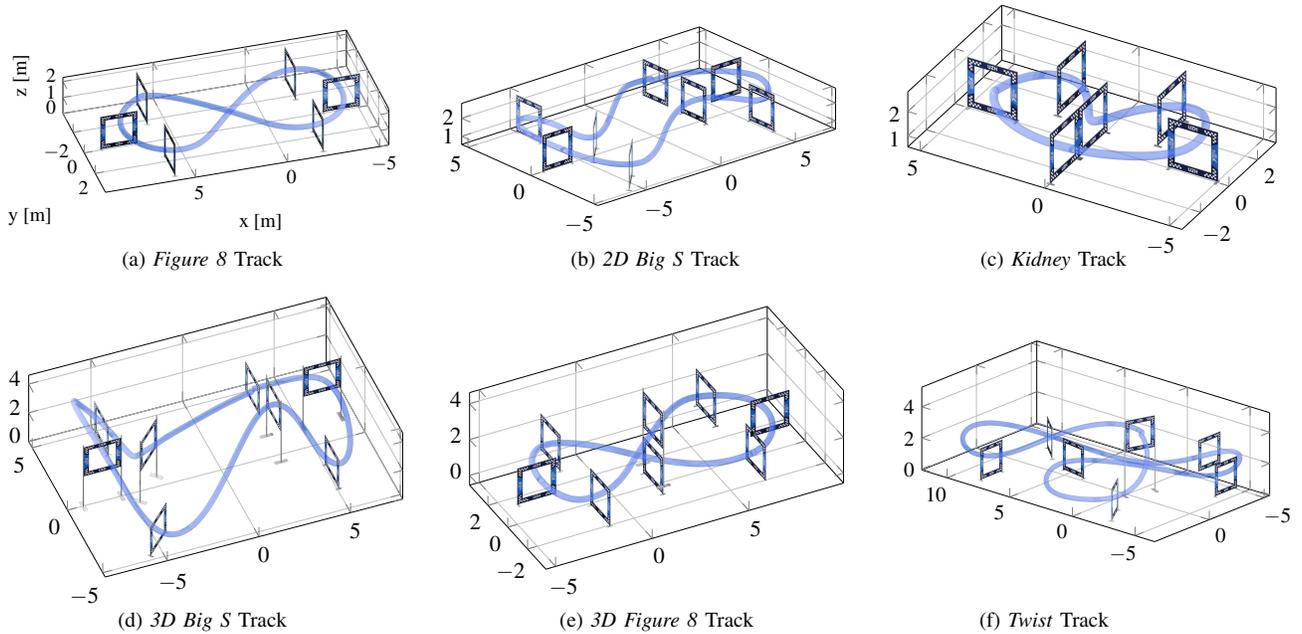
\begin{figure*}[ht!]
\centering
\small
\tikzstyle{every node}=[font=\footnotesize]
\begin{tikzpicture}
\node [inner sep=-10pt, outer sep=2pt] (img1) at (100,0) 
{\includegraphics[width=0.3\linewidth]{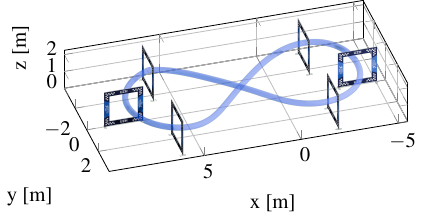}};
\node [inner sep=0pt, outer sep=20pt, right=0cm of img1] (img2) 
{\includegraphics[width=0.3\linewidth]{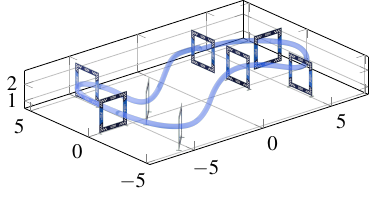}};
\node [inner sep=0pt, outer sep=0pt, right=-0.1cm of img2] (img3) 
{\includegraphics[width=0.3\linewidth]{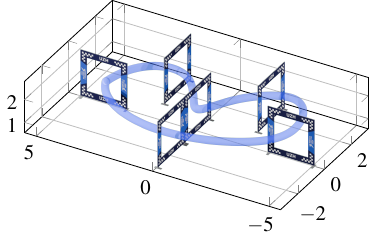}};

\node [inner sep=-0pt, outer sep=5pt, below=0.3cm of img1] (text1)
{(a) \textit{Figure 8} Track};
\node [inner sep=0pt, outer sep=10pt, right=3.3cm of text1] (text2) 
{(b) \textit{2D Big S} Track};
\node [inner sep=0pt, outer sep=10pt, right=2.5cm of text2] (text3) 
{(c) \textit{Kidney} Track};

\node [inner sep=-0pt, outer sep=5pt, below=0cm of text1] (img4)
{\includegraphics[width=0.3\linewidth]{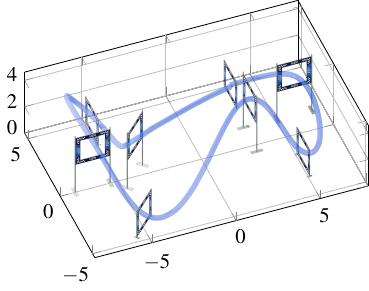}};
\node [inner sep=0pt, outer sep=10pt, right=0cm of img4] (img5) 
{\includegraphics[width=0.3\linewidth]{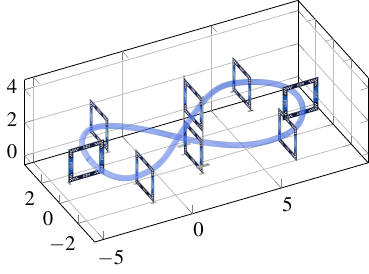}};
\node [inner sep=0pt, outer sep=10pt, right=0cm of img5] (img6) 
{\includegraphics[width=0.3\linewidth]{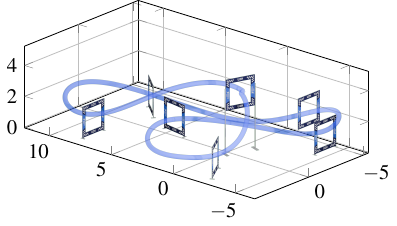}};

\node [inner sep=-0pt, outer sep=5pt, below=4.20cm of text1] (text4)
{(d) \textit{3D Big S} Track};
\node [inner sep=0pt, outer sep=10pt, right=3.1cm of text4] (text5) 
{(e) \textit{3D Figure 8} Track};
\node [inner sep=0pt, outer sep=10pt, right=2.3cm of text5] (text6) 
{(f) \textit{Twist} Track};
\end{tikzpicture}
\caption{Visualization of the drone racing tracks used for the experiments, each characterized by varying levels of complexity. All the tracks maintain a consistent size scale, spanning widths from 8 meters to 16 meters.}
\vspace{-0.3cm}
\label{fig:track}
\end{figure*}

Another key component of our framework is the development of a metric that precisely measures the effectiveness of the environment policy's actions on the performance of the racing agent.
In our work, we propose a reward for the environment policy grounded in the \textit{relative ranking} of the performance of the racing policy in different environments.
Specifically, after each training phase, we rank all the parallel training environments according to the number of gates the agent successfully passed.
The higher the ranking number, the worse the policy performs.
The environment policy is then penalized for generating track layouts at the extremes of the ranking—those that are either too easy or too difficult.
Additionally, we reward race tracks that fall within an intermediate range, where the agent demonstrates a degree of success but has not fully mastered the track.
The reward is defined as follows:
\begin{equation}
\vspace{-0.8em}
r^{env}_t =
\begin{cases} 
R\frac{N_{\mathrm{env}} - \mathrm{rank}}{N_{\mathrm{env}-r_\mathrm{upper}}} & \text{if } rank > r_\mathrm{upper} \\
R & \text{if } rank \in [r_\mathrm{lower}, r_\mathrm{upper}] \\
R \frac{\mathrm{rank}}{r_\mathrm{lower}} & \text{if } rank < r_\mathrm{lower}
\end{cases}
\vspace{0.3em}
\end{equation}
where \( R \) represents a positive constant, \( r_\mathrm{lower} \) and \( r_\mathrm{upper} \) are the fixed thresholds determining the mid-range representing the tracks that are not too easy and not too hard, $N_{\mathrm{env}}$ represents the total number of parallel training environments. 
This formulation represents a smooth reward function that provides fine-grained feedback on the environment policy for each generated training track.

\subsubsection{Environment Shaping}
To accelerate data collection, we train our environment agent and the racing agent using multiple parallel simulation environments.
To handle varying numbers of environments, the environment policy independently updates the track layout for each environment. 
Since all environments initially share the same track layout, the racing policy experiences a gradual increase in layout complexity, which facilitates smoother learning during the early stages of training. 
We use the notation $\mathrm{Env}_i \gets \mathrm{Env}_{i-1} \oplus a_{i-1}$ to represent changes in the individual environment. 
For each new environment, the environment policy is trained for every $N_{\text{freq}}$ racing policy update, where $N_{\text{freq}}$ determines how much more frequent should $\pi_\mathrm{racing}$ than $\pi_\mathrm{env}$.
Since $N_{\text{freq}}$ is typically much greater than 1 (i.e., $N_{\text{freq}} \gg 1$), the environment policy is updated less frequently than the racing policy.
To handle this, we choose the off-policy RL algorithm Soft Actor-Critic (SAC)~\cite{SAC} to update the parameters of the environment policy. 
The complete framework containing the environment policy and the racing policy during the training is presented in Algorithm~\ref{alg1}.
\begin{algorithm}[ht]
\begin{algorithmic}
\STATE \textbf{Initialize:} $\pi_{\mathrm{agent}}$, $\pi_{\mathrm{env}}$, and gate layout $\mathrm{Env}_0$
\FOR {$i\gets1, 2, ..., N_\mathrm{epoch}$}
    \STATE \textbf{Step 1: Generate New Environments}
    \IF {$i==1$}
    \STATE $a_0 \gets \text{random sample from the action space of } \pi_{\text{env}}$
        \STATE $\mathrm{Env}_1\gets \mathrm{Env}_0\oplus a_0$ 
    \ELSE
        \STATE $\mathrm{Env}_i\gets \mathrm{Env}_{i- 1}\oplus\pi_{\mathrm{env}}(\bm{s}_\mathrm{gates}, \bm{e}_\mathrm{racing})$
    \ENDIF
    \STATE \textbf{Step 2: Train racing policy}
    \FOR{$j\gets1, 2, ..., N_\mathrm{freq}$}
    \STATE Train $\pi_{\mathrm{agent}}$ on updated $\mathrm{Env}_{i}$
    \ENDFOR
    \STATE \textbf{Step 3: Evaluate and update environment policy}
    \STATE Evaluate agent performance across tracks $\mathrm{Env}_i$
    \STATE Compute relative ranking reward
    \STATE Update environment policy $\pi_{\mathrm{env}}$
\ENDFOR
\end{algorithmic}
\caption{Adaptive Environment Shaping}
\label{alg1}
\end{algorithm}
\vspace{-1ex}
\section{Experiments}
In this section, we first introduce the detailed experimental setups in Sec.~\ref{sec:expsetup}, followed by a discussion of the baseline approaches used for comparison in Sec.~\ref{sec:baseline}.
Our experiments aim to address the following key research questions:
(i) How does our approach generalize to unseen race tracks? 
(ii) How does our environment policy perform on different training configurations?
(iii) Can our generalist racing policy even fly in dynamic race tracks with moving gates?
(iv) Does our policy transfer to the real world?

\subsection{Experimental Setup}
\label{sec:expsetup}

\begin{table*}[ht!]
    \setlength{\tabcolsep}{5pt} 
    \renewcommand{\arraystretch}{1.2} 
    \begin{tabularx}{\textwidth}{>{\centering\arraybackslash}p{0.08\textwidth} >{\centering\arraybackslash}p{0.08\textwidth} *{2}{>{\centering\arraybackslash}p{0.06\textwidth}}|*{8}{>{\centering\arraybackslash}p{0.058\textwidth}}}
        \toprule \addlinespace[6pt]
        \multirow{3}*{\textbf{Track Type}} & \multirow{3}*{\textbf{Track Name}} & \multicolumn{10}{c}{\textbf{Methods}} \\
        \cmidrule(lr){3-12} 
        \multicolumn{2}{c}{} & \multicolumn{2}{c|}{\textit{Single-track RL}} & \multicolumn{2}{c}{\textit{RL w/o curriculum}} & \multicolumn{2}{c}{\textit{Particle Filter}} & \multicolumn{2}{c}{\textit{Domain Randomization}} & \multicolumn{2}{c}{\textit{Ours}} \\
        \cmidrule(lr){3-4} \cmidrule(lr){5-6} \cmidrule(lr){7-8} \cmidrule(lr){9-10} \cmidrule(lr){11-12}
        \multicolumn{2}{c}{} & SR [\%] & LT [s] & SR [\%] & LT [s] & SR [\%] & LT [s] & SR [\%] & LT [s] & SR [\%] & LT [s] \\
        \midrule
        \grayrow
        \multirow{3}*{\textbf{2D}} 
        & \textit{Figure 8}       & 100.00 & 4.263 & 0.00 & - & \bfseries 100.00 & 5.128 & \bfseries 100.00 & 6.205 & \bfseries 100.00 & \bfseries 4.746 \\
        & \textit{Kidney}         & 100.00 & 4.260 & 0.00 & - & 0.00 & - & \bfseries 100.00 & 4.984 & \bfseries 100.00 & \bfseries 4.943 \\
        \grayrow & \textit{Big S} & 100.00 & 9.245 & 0.00 & - & 0.00 & - & 0.00 & - & \bfseries 100.00 & \bfseries 7.513 \\
        \midrule
        \grayrow
        \multirow{3}*{\textbf{3D}} 
        & 3D \textit{Figure 8}    & 100.00 & 5.010 & 0.00 & - &\bfseries 100.00 &\bfseries 5.654 & 0.00 & - &  \bfseries 100.00 &  5.856 \\
        & 3D \textit{Big S}       & 100.00 & 10.187 & 0.00 & - & 0.00 & - & 0.00 & - & \bfseries 100.00 & \bfseries 9.761 \\
        \grayrow & \textit{Twist} & 100.00 & 7.444 & 0.00 & - & 0.00 & - & 0.00 & - & \bfseries 100.00 & \bfseries 10.199 \\
        \bottomrule
    \end{tabularx}
    \caption{We compare the success rate (SR) and lap time (LT) of our method against four baselines. Six different unseen racetracks are evaluated in a realistic BEM simulation~\cite{bauersfeld2021neurobem}, including three 2D tracks and three 3D tracks. Here apart from the our and baseline approaches, we also include the \textit{Single-track RL} as a reference, which is a vanilla PPO agent trained and tested on individual track.}
    \vspace{-2em}
    \label{Diff}
\end{table*}
Our training framework is built on the Flightmare simulator~\cite{song2020flightmare}, with reward functions and PPO hyperparameters aligned with methods from previous research~\cite{kaufmann23champion}.
For our environment policy, we implemented a vectorized Gym~\cite{GYM} environment with the same number of environments as the agent, using a total of 100 environments. 
The environment policy uses the SAC implementation from Stable Baselines 3~\cite{stable-baselines3}, with fine-tuned parameters for our task setting.
Specifically, we set the gradient steps of SAC to 100 to accelerate the training of the environment policy and the entropy coefficient to 1.0 for better exploration. 

During the training process, the racing policy is updated at every iteration, while the environment policy is updated every 100 iterations. 
We sometimes encounter environments getting stuck on infeasible tracks during training. To address this, if the evaluation SR stays at zero for three consecutive runs, we reset the environment to the initial race track.
We selected an 8-gate oval track for the initial layout due to its simplicity, making it easier for the racing policy to learn and adapt quickly.
For the environment policy's action space, we constrained the gate movements as follows: along the x and y axes, the range is $[-1.0, 1.0]$$\si{\meter}$; along the z-axis, the range is $[-0.2, 0.2]$$\si{\meter}$, and for the yaw, the range is \([- \frac{\pi}{30}, \frac{\pi}{30}]\) $\si{\radian}$. 
For the relative ranking range, we set $r_{lower}$ to the 50th percentile of all environments and $r_{upper}$ to the 90th percentile.
These hyperparameters are empirically found to lead to the best performance during the ablation experiments.
Our method is trained for 800M data samples, which is equivalent to 12 hours of training using an NVIDIA TITAN Xp GPU.

To evaluate the performance of our approach, we tested the racing policy on several fixed racetracks with varying difficulties and complexities.
We selected three 2D tracks, where all gates are positioned at the same height along the z-axis, including \textit{Figure 8}, \textit{Kidney}, and \textit{2D Big S}. Additionally, we chose three 3D tracks, where the gates have varying heights along the z-axis, including \textit{3D Figure 8}, \textit{3D Big S}, \textit{and Twist}, as shown in Fig.~\ref{fig:track}.
All of these tracks are significantly different than the initial training track, with most having a different number of gates than during training. 
For the evaluation, we primarily used two metrics: success rate (SR \%) and lap time (LT [\si{\second}]).
The success rate is calculated as the ratio of successful runs, where the drone passes all the gates without crashing, over the total number of trials.
Lap time refers to the drone's total time to successfully complete a race track. 
These metrics are commonly used in drone racing and are essential in evaluating whether the policy enables fast and stable flight performance~\cite{geles2024demonstrating, xing2024bootstrapping}. 

\subsection{Baselines.}
\label{sec:baseline}
To evaluate the generalization ability of our proposed methods, we compared them with three baseline methods:
(i) RL policy without curriculum: This method shares the same initial environment setup as ours but does not use a curriculum for training.
(ii) Domain randomization from~\cite{song2021autonomous}: In this approach, the initial environment is the same as in our method. However, environment policy actions are sampled randomly within the action space; we use the identical configurations from~\cite{song2021autonomous}.
(iii) Particle Filter Curriculum from~\cite{lee2020learning}: This method also uses the same initial environment as ours but relies on a particle filter to sample environments.

\textbf{How does our approach generalize to unseen race tracks?}
We first evaluated our method on six unseen tracks to assess its generalization ability across tracks with varying numbers of gates and different layouts compared to those used during training. 
Table~\ref{Diff} presents the performance results.
As can be observed, the reference single-track RL policies on the left side demonstrate the best performance in most experiments. 
This is because these policies overfit the specific track, allowing them to optimize and perform well in that particular environment.
Consequently, the results from the RL method without a curriculum show that directly training on one track without a curriculum and testing on different tracks is not successful. 

Additionally, we found that a simple curriculum has a limited effect on improving the racing policy's generalization ability. 
Methods such as domain randomization and particle filters can enhance the agent's generalization ability, allowing it to fly on some simple unseen tracks, such as \textit{Figure 8} or \textit{Kidney} tracks.
However, since these methods do not take into account the continuous improvement of the agent's policy during training or the evolution of the environment, they fail to perform well in rather complicated unseen tracks, e.g., in \textit{2D Big S} or \textit{Twist} tracks. 
Only our proposed method demonstrated stable and successful flight across all six unseen tracks.
Additionally, we observed that for most 2D and 3D tracks (excluding the \textit{Twist} track), the policy's performance (lap time) remains close to that of the \textit{single-track RL policy}, within a 1-second difference.
This difference can be explained by the increased generalization of our policy, which did not memorize a single specific track.
This confirms that our training approach maintains strong task performance while enhancing generalization. 

\bigskip

\textbf{How does our environment policy perform on different training configurations?}
\begin{figure}[t]
    \centering
 
\includegraphics[width=0.9\linewidth]{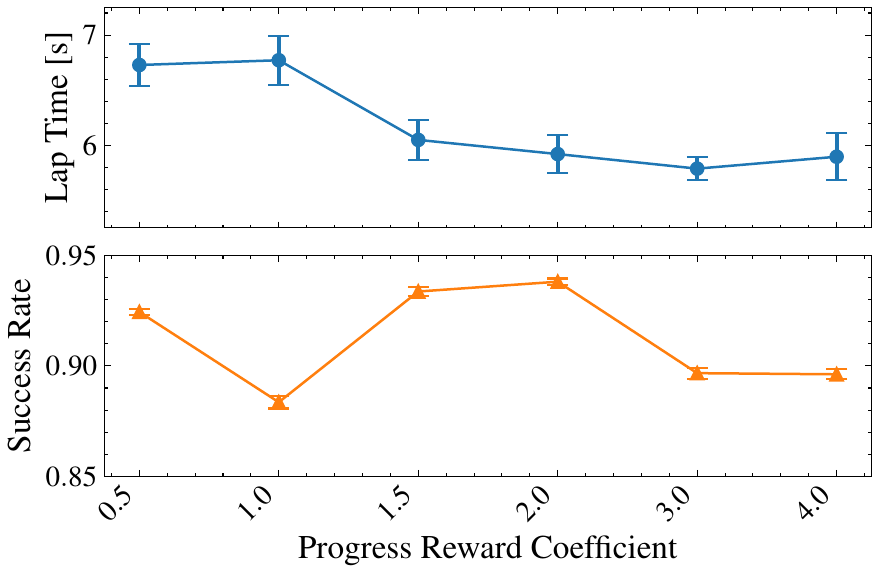}

    \caption{Ablation study on the progress reward. Due to the fluctuations in evaluation results across different iterations, to fairly compare the performance of different coefficients, we take the average and variance of the success rate and lap time after the model has stabilized for comparison. }
    \label{fig:reward}
    \vspace{-4ex}
\end{figure}
To assess the impact of different configurations on the performance of our environment policy in racing policy training, we performed an ablation study.
Within the same parameter setting for the environment policy, we vary the progress reward coefficients $\alpha_1$ from Equation~\ref{eq:reward} of the racing policy.
To the variance in the results, we tested the different configurations on 100 randomly generated tracks, which were manually filtered for feasibility.
As shown in Fig.~\ref{fig:reward}, within a large range of parameters, increasing the progress reward coefficient helps the agent learn to fly faster, as evidenced by the decreasing lap times on specific tracks. 
At the same time, we observe no significant decrease in generalization, as the success rate remains fairly stable. 
This indicates the robustness of our framework, where a dedicated human-defined curriculum usually needs to be re-designed for different speed ranges.

\textbf{Can our generalist racing policy fly in dynamic race tracks with moving gates?}
To evaluate the generalization of our method, we further test the racing policy by deploying it on a dynamic track. 
In this scenario, several gates move at a constant speed in predefined directions.
Notably, this dynamic setting was never included in the training configurations of the racing policy.
As the track continuously changes, the racing policy's observations evolve over time. 
In this setting, only a highly robust control ability can finish the lap and prevent crashes.
In this experiment, we chose the \textit{Figure 8} track, where the third and fourth gates move dynamically along their y-axis at a constant speed of 0.6\si{\meter\per\second} within the ranges of [-0.5, 0.5]\;\si{\meter}, [-1.0, 1.0]\;\si{\meter}, and [-2.0, 2.0]\;\si{\meter}.
As can be observed in Table~\ref{Diff1}, we can see that the Domain Randomization~\cite{song2021autonomous} and Particle Filter~\cite{lee2020learning} methods still achieve partial success when the gates move within the range of [-0.5, 0.5]\;\si{\meter}.
However, both methods fail when the range increases to [-1.0, 1.0]\;\si{\meter}. 
In contrast, our method can handle perturbations up to [-2.0, 2.0]\;\si{\meter}. 
This further demonstrates our method's superior generalization ability, enabling it to adapt its actions as observations change and maintain stable performance even in the face of significant dynamic variations.
Also, we test the current limit of our approach, where we further increase the gate moving speed to $0.75\si{\meter\per\second}$.
In this setting, our policy can still achieve more than $30\%$ success rate.

\textbf{Does our policy transfer in the real world?}
To validate the effectiveness of our proposed method, we conducted tests in real-world conditions.
We used the Agilicious quadrotor platform~\cite{foehn2022agilicious} with precise state estimation provided by a VICON motion capture system, ensuring accurate inputs for the policy.
The BetaFlight2 firmware was employed for low-level control to execute the collective thrusts and body rate commands.
We performed nine laps on each of the six tracks (Fig.~\ref{fig:track}), demonstrating that our single racing policy can successfully navigate these previously unseen tracks in the real world with a success rate of 100$\%$, as shown on the right side of Fig.~\ref{fig:overview}.
For further details, we invite readers to view the supplementary video.

\begin{table}[t]
    \setlength{\tabcolsep}{3pt} 
    \renewcommand{\arraystretch}{1.2} 
    \begin{tabularx}{\columnwidth}{>{\centering\arraybackslash}p{0.08\textwidth}>{\centering\arraybackslash}p{0.16\textwidth} *{2}{>{\centering\arraybackslash}p{0.095\textwidth}}}
        \toprule
        \textbf{Range} [\si{\meter}] & \textbf{Methods} & SR [\%] & LT [s] \\
        \midrule
        \grayrow
        \multirow{3}*{Static} & Single-track RL & 100.00 & 4.263\\
               & Particle Filter & 100.00 & 5.128\\
        \grayrow
               & Domain Randomization & 100.00 & 6.205\\
               & \bfseries Ours & \bfseries 100.00 & \bfseries 4.746  \\
        \cmidrule(lr){1-4} 
        \grayrow
        \multirow{3}*{$[-0.5, 0.5]$} & Particle Filter & 54.69 & 5.512\\
               & Domain Randomization & 93.75 & 6.152\\
        \grayrow
               & \bfseries Ours & \bfseries 100.00 & \bfseries 5.388  \\
        \addlinespace[3pt]
        \cmidrule(lr){1-4} 
         & Particle Filter & 0.00 & -\\
           \grayrow    $[-1, 1]$& Domain Randomization & 0.00 & -\\
               & \bfseries Ours & \bfseries 100.00 & \bfseries 5.276  \\
        \addlinespace[3pt]
        \cmidrule(lr){1-4} 
        \grayrow & Particle Filter & 0.00 & -\\
               & Domain Randomization & 0.00 & -\\
               \grayrow $[-2, 2]$
               & \bfseries Ours (0.6m/s)& \bfseries 100.00 & \bfseries 5.329  \\
               & \bfseries Ours (0.75m/s) & \bfseries 31.25 & \bfseries 5.067  \\
        \addlinespace[3pt]
        \bottomrule
    \end{tabularx}
     \caption{Results of flying on a dynamic \textit{Figure 8} racetrack. We compare the success rate (SR) and lap time (LT) of different methods.}
         \label{Diff1}
    \vspace{-1.8em}
\end{table}

\section{Conclusion}
In this work, we proposed an adaptive environment-shaping framework enabling, for the first time, a learned policy to race on unseen and dynamic tracks.
Our method works by leveraging a secondary environment policy that shapes the training environments for a drone racing policy.
Using our novel relative ranking reward, our environment policy can generate track layouts that are challenging but feasible based on the racing agents' performance.
One single drone racing policy trained with our framework can race in various race tracks with different complexity with $100\%$ success rate, whereas the state-of-art curriculum learning approach mostly cannot fly.
Furthermore, we validated the policy's generalization ability by racing on a track with moving gates, where existing methods performed significantly worse at adapting to the fast-changing gate positions.
We believe our work represents a significant advancement in enabling agile robots to achieve greater generalization and robustness in complex, dynamic, and open environments.
\clearpage
\bibliographystyle{ieeetr}
\bibliography{icra}

\end{document}